\documentclass[sigconf]{acmart}
\AtBeginDocument{%
  \providecommand\BibTeX{{%
    \normalfont B\kern-0.5em{\scshape i\kern-0.25em b}\kern-0.8em\TeX}}}

\acmConference[ACM MM'22]{ACM Multimedia conference}{October 2022}{Lisbon Portugal}
\acmYear{2022}
\copyrightyear{2022}

\usepackage{hyperref}
\newcommand{\etal}{\textit{et al.~}}

\begin{document}

\title{IVT: An End-to-End Instance-guided Video Transformer for 3D Pose Estimation}

\author{Zhongwei Qiu}
\authornote{Equal Contribution}
\affiliation{%
  \institution{University of Science and Technology Beijing}
  \country{}}
\email{qiuzhongwei@xs.ustb.edu.cn}

\author{Qiansheng Yang}
\authornotemark[1]
\affiliation{%
  \institution{Baidu Inc.}
  \country{}}
\email{yangqiansheng@baidu.com}

\author{Jian Wang}
\affiliation{%
  \institution{Baidu Inc.}
  \country{}}
\email{wangjian33@baidu.com}

\author{Dongmei Fu}
\affiliation{%
\institution{University of Science and Technology Beijing}
  \country{}}
\email{fdm_ustb@ustb.edu.cn}

\begin{abstract}
Video 3D human pose estimation aims to localize the 3D coordinates of human joints from videos. Recent transformer-based approaches focus on capturing the spatiotemporal information from sequential 2D poses, which cannot model the contextual depth feature effectively since the visual depth features are lost in the step of 2D pose estimation. In this paper, we simplify the paradigm into an end-to-end framework, Instance-guided Video Transformer (IVT), which enables learning spatiotemporal contextual depth information from visual features effectively and predicts 3D poses directly from video frames. In particular, we firstly formulate video frames as a series of instance-guided tokens and each token is in charge of predicting the 3D pose of a human instance. These tokens contain body structure information since they are extracted by the guidance of joint offsets from the human center to the corresponding body joints. Then, these tokens are sent into IVT for learning spatiotemporal contextual depth. In addition, we propose a cross-scale instance-guided attention mechanism to handle the variational scales among multiple persons. Finally, the 3D poses of each person are decoded from instance-guided tokens by coordinate regression. Experiments on three widely-used 3D pose estimation benchmarks show that the proposed IVT achieves state-of-the-art performances.
\end{abstract}



\maketitle

\section{Introduction}
3D human pose estimation aims to localize the 3D joints of person(s) from monocular images or videos. As a fundamental computer vision task, it has a lot of applications, including action recognition~\cite{liu2020disentangling}, human-robot interaction detection~\cite{li2020detailed}, and virtual reality~\cite{parger2021unoc}, etc. Unfortunately, estimating 3D human poses from monocular 2D images or videos is very challenging because of the lack of depth information.

\begin{figure}[!t]
\centering
\includegraphics[width=\columnwidth]{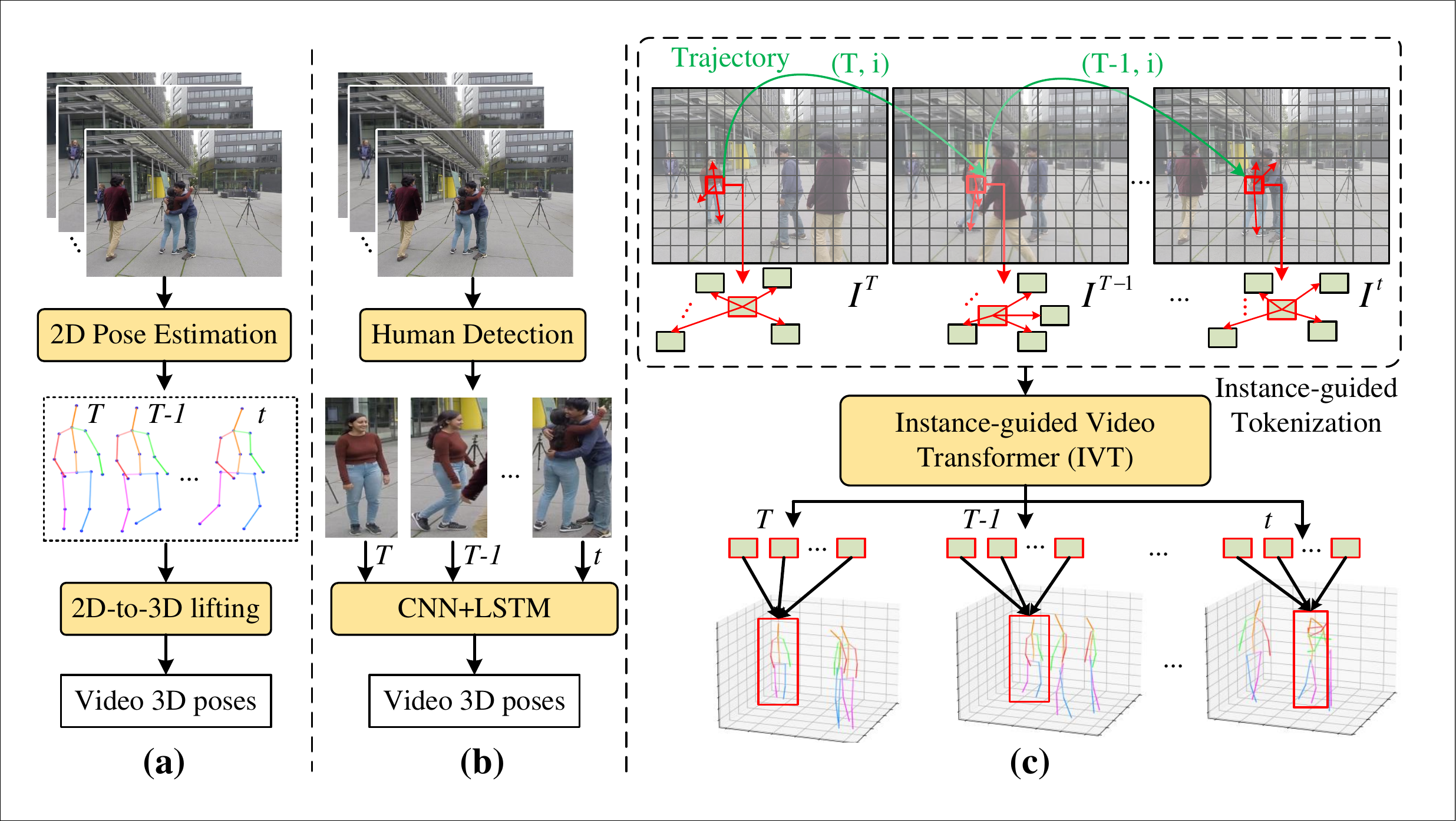}
\vspace{-0.5cm}

\caption{Comparison of (a) The pipeline of 2D-to-3D video pose lifting~\cite{cai2019exploiting,zheng20213d}, (b) Recurrent structure based methods~\cite{kocabas2020vibe,choi2021beyond} for video 3D pose estimation, (c) Our end-to-end Instance-guided Video Transformer (IVT). IVT is a single-stage framework while others are two-stage methods.}
\label{fig:fig1}
\end{figure}

To tackle this problem, some image-based approaches learn depth information from image feature by depth map supervision~\cite{wang2020hmor} or 3D heatmap supervision~\cite{moon2019camera,sun2021monocular}, while other image-based approaches~\cite{yang20183d,zeng2020srnet} firstly estimate 2D pose from image, and then lift 2D pose to 3D pose. However, the depth information implied in a single image is still limited. Compare to a single image, video can provide more motion cues which are quite helpful for the inference of depth. Thus, video-based approaches~\cite{sun2019human,arnab2019exploiting,cai2019exploiting,cheng20203d,choi2021beyond,zheng20213d} are developed rapidly in recent years. And we also focus on the video-based method in this paper.

By the benefit of the amazing performance of modern 2D pose estimators, many video-based approaches follow the 2D-to-3D lifting paradigm as Figure \ref{fig:fig1} (a), and exploit spatial and temporal modeling methods to improve the performance of 3D pose estimation. 
\cite{liu2020attention} and \cite{hossain2018exploiting} apply convolution neural network (CNN) and recurrent neural network (RNN) respectively to model temporal dependency among the sequential 2D poses. 
But limited by the formulation of CNN and RNN, they are not good at capturing the long-range dependency in both spatial and temporal dimensions. 
To alleviate this situation, some works~\cite{cai2019exploiting,wang2020motion,cheng20203d} introduce graph neural network (GNN) to exploit spatial-temporal information between keypoints, which can capture both short-term and long-term dependency by setting an appropriate adjacent matrix. Besides GNN, other works~\cite{lin2021end,zheng20213d} use transformer to get more representative features from pose sequence, and also improve the performance of video 3D pose estimation significantly.

Despite the significant progress achieved by the above GNN or transformer-based methods, they still did not break out of the 2D-to-3D lifting paradigm. This paradigm only considers the structure of the 2D pose for depth estimation, while ignoring the contextual depth information contained in the semantic feature of video frames, since the semantic feature has been dropped at the 2D pose prediction stage. But we suppose that the context depth information embedded in the semantic feature is more effective than the 2D pose structure for 3D pose estimation.

As shown in Figure \ref{fig:fig1} (b), some video-based approaches~\cite{sun2019human,choi2021beyond} with the recurrent neural network, firstly conduct human detection and then predict 3D pose directly from the cropped video patches, which can be regarded as exploiting contextual depth information from the semantic feature of video frames to some extent. But they apply the recurrent neural network to exchange features between different video frames for only temporal modeling, which is not effective compared with GNN or transformer-based spatial-temporal modeling methods mentioned above. Besides, the inputs of cropped patches bring a new problem of keypoints feature alignment.

To handle the above problems, we simplify the video 3D pose estimation into an end-to-end transformer-based framework as shown in Figure \ref{fig:fig1} (c), which aims to make full usage of the spatial-temporal depth feature and predicts 3D pose directly from video frames. In order to capture the effective contextual depth information and reduce the computational burden introduced by conducting self-attention on the dense semantic feature of video frames, we propose an Instance-guided Video Transformer to model the spatial-temporal depth information by the guidance of human instance. 

Firstly, Instance-guided Video Transformer (IVT) introduces a series of instance-guided visual tokens and each token is capable for predicting 3D pose of an instance. These tokens are contructed by aggregating features from related spatial points by the guidance of a set of learned 2D offsets from human center to the corresponding human joints. This mechanism enables each token can capture the whole body information. For a query token, the attention is computed on both spatial and temporal dimensions to exchange the context depth information. Furthermore, we propose a cross-scale instance-guided attention mechanism to handle the variational scales among multiple persons. As a result, IVT enables effective spatial-temporal depth feature exchange and brings significant improvement to video 3D pose estimation.

In summary, IVT is a simple and unified framework that is suitable for both single-person and multi-person video 3D pose estimation tasks. And to the best of our knowledge, for multi-person video 3D pose estimation, IVT is the first end-to-end method that leverages transformer to directly capture multi-person depth information in the video. Our contributions can be summarized as follows:
\begin{itemize}
    \item We propose a novel end-to-end transformer-based framework for both single-person and multi-person video 3D pose estimation, called instance-guided video transformer (IVT).
    \item We design a novel instance-guided attention mechanism to enable effective spatial-temporal depth information learning in videos.
    \item IVT achieves new state-of-the-art results on three widely-used video 3D pose estimation benchmarks, Human3.6M, 3DPW, and CMU Panoptic datasets.
\end{itemize}

\section{Related Work}
\subsection{Image-based 3D Pose Estimation}
The image-based multi-person 3D pose estimation methods can be mainly divided into two kinds of paradigms: top-down~\cite{yang20183d,moon2019camera,zeng2020srnet,lin2021end,sun2021monocular} and bottom-up approaches~\cite{zanfir2018deep,wang2020hmor,zhen2020smap}.

The top-down paradigm follows a pipeline of conducting human detection firstly and performing single-person 3D pose estimation later. For the single-person cases, they predict 3D poses by learning 3D heatmaps~\cite{moon2019camera}, or estimating 2D poses by 2D pose estimator~\cite{qiu2019learning,qiu2020dgcn} and lifting 2D poses to 3D poses~\cite{zeng2020srnet}. Typically, PoseNet~\cite{moon2019camera} predicts the root depths of each person at the stage of human detection, then estimates the 3D coordinates from 3D heatmaps. The bottom-up paradigm~\cite{zanfir2018deep,wang2020hmor,zhen2020smap} follows a pipeline of firstly estimating the 3D coordinates for each human joint in an image and then assigning them to different human instances. For example, MubyNet~\cite{zanfir2018deep} estimates keypoints and limb core at the same time and then integrates limb score to group keypoints into different persons. HMOR~\cite{wang2020hmor} propose hierarchical multi-person ordinal relations as an additional loss to help depth learning. However, the image-based approaches are not good at handling occlusion cases as the video-based approaches.

\subsection{Video-based 3D Pose Estimation}
Video-based multi-person 3D pose estimation aims to capture temporal information for 3D pose estimation. The ways of extracting temporal information can be divided into two categories: based on image visual features~\cite{sun2019human,kocabas2020vibe,choi2021beyond} and based on 2D poses~\cite{arnab2019exploiting,cai2019exploiting,zheng20213d}.

The methods~\cite{sun2019human,kocabas2020vibe,choi2021beyond} based on image visual features usually crop the human features according to human bounding boxes, and then use 3D convolution or recurrent neural network to extract the temporal information from these cropped sequences features. Sun \etal~\cite{sun2019human} propose a skeleton-disentangling framework to separate 3D human pose and shape estimation into spatial and temporal dimensions. TCMR~\cite{choi2021beyond} uses ResNet to extract visual features from video frames, then captures temporal information on these deep features by the recurrent neural network. However, these methods essentially conduct single-person video 3D pose estimation, which brings a new problem of feature alignment due to crop images. 

The methods~\cite{arnab2019exploiting,cai2019exploiting,zheng20213d} based on 2D coordinates usually estimate a sequence of 2D poses at first, then lift 2D coordinates sequence to 3D pose by a temporal lifting network. Typically, Cai \etal~\cite{cai2019exploiting} exploit spatial-temporal relationships for 3D pose estimation via Graph Convolutional Networks (GCN). Zheng~\etal~\cite{zheng20213d} propose PoseFormer, a spatial-temporal transformer network to capture the spatial-temporal information among human joints. However, these methods cannot capture truly depth features from visual images since the depth feature is lost in the stage of 2D human pose estimation. Besides, these methods disassemble the multi-person video task into a single-person video task. Thus, the depth information between different persons can not be captured. In this paper, we tackle the problems and build an end-to-end multi-person video 3D pose estimation framework.

\subsection{Transformer in Human Pose Estimation}
Recently, the transformer-based approaches~\cite{yang2021transpose,li2021pose,mao2021tfpose,lin2021end,zheng20213d,huang2021unifying} have been proposed to improve the long-term modeling capabilities of sequence for human pose estimation. TransPose~\cite{yang2021transpose} and TFPose~\cite{mao2021tfpose} formulate human joints as visual tokens and capture the relationship between human joints by self-attention. METRO~\cite{li2021pose} and PRTR~\cite{li2021pose} exploit the end-to-end transformer-based pose estimation network. PoseFormer~\cite{zheng20213d} explores the spatial-temporal attention mechanism for 3D pose estimation. However, the PoseFormer didn't study the attention on real depth features from images since it lifts 3D poses from a sequence of 2D poses. Moreover, the existing transformer-based pose estimation methods are designed for single-person pose estimation, which limits their applications. In this paper, we study an end-to-end multi-person 3D pose estimation framework and explore to extract the relationship between multi-person joints in both spatial and temporal dimensions.

\section{Method}

\begin{figure*}[!t]
\centering
\includegraphics[width=\textwidth]{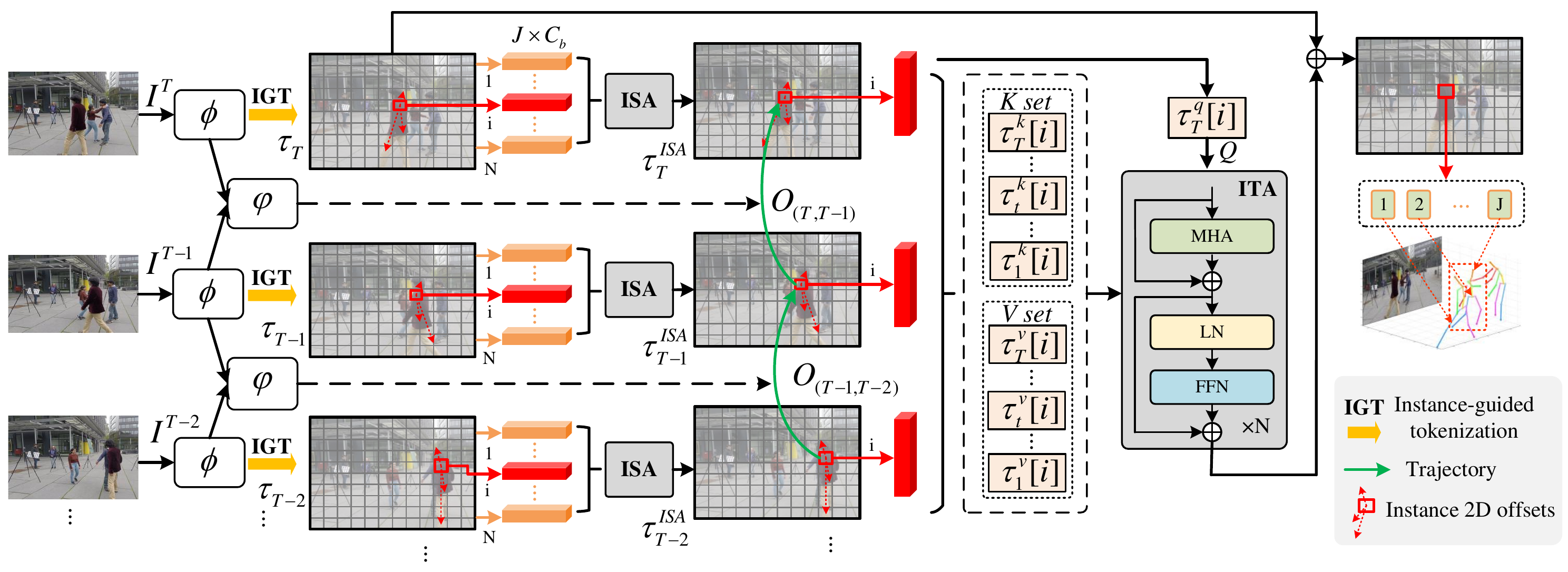}
\vspace{-0.8cm}
\caption{The overview of Instance-guided Video Transformer (IVT), which includes instance-guided tokenization (IGT), instance-guided spatial attention (ISA), and instance-guided temporal attention (ITA). Given video frames of height $H$ and width $W$, deep features are extracted by embedding network $\phi (\cdot)$, and are further used to estimate trajectory motions $O$ by network $\varphi(\cdot)$. The keypoints features are extracted from deep features according to the trajectory and instance 2D offsets and to further learn instance-aware tokens of shape $J\times C_b$ by IGT. For a query token $\tau^q_{T}[i]$ at the $i^{th}$ block in $T^{th}$ frame, ISA is computed in each frame and ITA is computed among temporal frames to capture depth information. Each token in the final layer of IVT outputs the 3D coordinates of a person. MHA, LN, and FFN denote multi-head attention, layer norm, and feed-forward network, respectively. $\oplus$ means element-wise addition. $J$ represents the joints number. N represents token numbers in each frame.}
\label{fig:framework}
\end{figure*}

In this section, we elaborate on the detail of the proposed Instance-guided Video Transformer (IVT). The framework of IVT is shown in Figure \ref{fig:framework}. Given a sequence of video frames $I=\{I_t\ |\ t \in [1, T]\}$, IVT firstly extracts deep features by a backbone network, which are used to learn instance 2D offsets (from body center to $J$ keypoints) and temporal feature motion (trajectory).
Then, for each frame, the deep features are organized as visual tokens with the guidance of instance 2D offsets, called instance-guided tokens. And each instance-guided token is in charge of predicting the 3D pose of its corresponding instance by the assistant of the aggregated whole body information. This process is denoted as Instance-Guided Tokenization (IGT).
After conducting IGT, instance-guided tokens are sent into a video transformer to capture context depth information in both spatial and temporal dimensions. 
Finally, the outputted visual tokens are further used to decode 3D poses for persons detected from a human center heatmap.

\subsection{Formulation of Transformer Module}

First of all, we review the formulation of the basic transformer module. Given query matrix $\mathcal{Q}$, key matrix $\mathcal{K}$, value matrix $\mathcal{V}$, where the first dimension of them is sample dimension while the second is feature dimension, a typical attention formulation $A(\mathcal{Q}, \mathcal{K}, \mathcal{V})$ can be expressed as:
\begin{equation}
    A(\mathcal{Q}, \mathcal{K}, \mathcal{V}) = softmax(\frac{\mathcal{Q} \cdot \mathcal{K}^T}{\sqrt{d}}) \cdot \mathcal{V}
\end{equation}
where $d$ is the length of feature within these matrixs. If we split $\mathcal{Q}$, $\mathcal{K}$, $\mathcal{V}$ into $h$ heads via feature axis and conduct attention for each head, denoted as $\{\mathcal{Q}_1, ..., \mathcal{Q}_h\}$, $\{\mathcal{K}_1, ..., \mathcal{K}_h\}$, $\{\mathcal{V}_1, ..., \mathcal{V}_h\}$, the Multi-Head Attention $MHA(\mathcal{Q}, \mathcal{K}, \mathcal{V})$ can be formulated as:
\begin{equation}
\begin{aligned}
    MHA(\mathcal{Q}, \mathcal{K}, \mathcal{V}) &= P(Concat_c(head_1, ..., head_h))\\
    head_i &= A(\mathcal{Q}_i, \mathcal{K}_i, \mathcal{V}_i), i \in [1, h]
\end{aligned}
\end{equation}
where $\mathbf{P}$ is linear projection function, $Concat_c$ means concatenating matrixs along feature axis. Specially, when $\mathcal{Q}$, $\mathcal{K}$, $\mathcal{V}$ are derived from a same input matrix $\mathcal{X}$, we can get Multi-Head Self-Attention $MHSA(\cdot)$ further, which is denoted as:
\begin{equation}
\begin{aligned}
    &MHSA(\mathcal{X}) = MHA(\mathcal{Q}, \mathcal{K}, \mathcal{V})\\
    &\mathcal{Q} = \mathbf{P}_q(\mathcal{X}), \mathcal{K} = \mathbf{P}_k(\mathcal{X}), \mathcal{V} = \mathbf{P}_v(\mathcal{X})
\end{aligned}
\end{equation}
where $\mathbf{P}_q$, $\mathbf{P}_k$, $\mathbf{P}_v$ are linear projection functions for generating $\mathcal{Q}$, $\mathcal{K}$, $\mathcal{V}$, respectively.

Then, two basic transformer modules used in this paper can be formulated as:
\begin{align}
    \mathcal{X}_{out} &= FFN(MHSA(\mathcal{X}_{in}))\\
    \mathcal{X}_{out} &= FFN(MHA(\mathcal{Q}, \mathcal{K}, \mathcal{V}))
\end{align}
and they are served for self-attention and cross-attention respectively. While, in the above equations, $FFN(\cdot)$ represents feed forward network, which consists of two linear layers. For simple expression, the layer norm and shortcut path are ignored here.

\subsection{Instance-guided Tokenization}

In this section, we introduce the process of generating instance-guided tokens for each video frame $I_t$, named Instance-Guided Tokenization (IGT).

We firstly extract deep feature $F_t$ for $I_t$ by a backbone network $\phi (\cdot)$, and the shape of $F_t$ is $C\times H\times W$. 
Then we split $F_t$ into $N$ feature blocks, denote as $B_t$, which has a shape of $N\times C_b$, where $C_b = C\times K\times K$ and $K$ represents the block size. 
To extract visual tokens, traditional vision transformers~\cite{dosovitskiy2020image,bertasius2021space} take each block as a token and capture the spatial and temporal relationships among these block tokens. Then, $N$ tokens are generated from features $B_t$, denoted as $\tau_t$. The shape of $\tau_t$ is $N\times C_t$, where $C_t = C_b$ is feature dim of tokens. However, this tokenization is not fine-grained to capture the context information from the human body to predict depth for 3D human pose estimation. 
Here, we introduce the instance-guided tokenization (IGT) approach. Different from traditional tokenization, IGT considers the features from whole body as well as the relationship between human joints when extracting visual tokens, which enables each token to encode the body structure of its corresponding human instance.

For the $i$th block in $B_t$, denoted as $B_t[i]$, the process of instance-guided tokenization includes two steps. Firstly, it gathers features from $J$ corresponding blocks of $B_t[i]$ with the guidance of instance 2D offsets. As shown in Figure \ref{fig:framework}, instance 2D offsets are the joint offsets from body center to $J$ joints, and these offsets are preserved in an offset map $M^{2D}_o$, which is predicted by several convolutional layers based on deep feature $F_t$. $M^{2D}_o$ contains the whole body information of each instance and indicates the feature locations of relative keypoints.
The gathered blocks are concatenated into one feature vector, denoted as $\widetilde{B}_t[i]$, and its length is $J\times C_b$. Secondly, a multi-head self-attention module is used to encode $\widetilde{B}_t[i]$ and generate instance-guided token $\tau_t[i]$. This enables feature exchange between different joints of a human instance and makes the token feature be aware of body structure, which improves the robustness of instance-guided token for predicting whole body 3D pose. Concretely, this self-attention can be formulated as:
\begin{equation}
\begin{aligned}
    f &= reshape(\widetilde{B}_t[i], (J, C_b))\\
    f &= FFN(MHSA(f))\\
    \tau_t[i] &= reshape(f, J\times C_b)
\end{aligned}
\end{equation}
where $MHSA$ and $FFN$ are multi-head self-attention module and feed forward module respectively. $reshape(f, shape)$ means reshaping the input data $f$ into target $shape$.
As a result, each generated instance-guided token $\tau_t[i]$ can encode the global context from a human instance.

Once the token feature $\tau_t$ for each video frame $I_t$ is extracted, we pass all the token features $\tau = \{\tau_t | t \in [1, T]\}$ of $T$ video frames to the instance-guided video transformer for 3D pose estimation, which will be elaborated in the next section.

\subsection{Instance-guided Video Transformer}

Instance-guided video transformer (IVT) takes the instance-guided tokens $\tau_t$ as inputs and conducts spatial-temporal attention to capture context depth information in both spatial and temporal dimensions. It can be divided into two sequential attention stages, Instance-guided Spatial Attention(ISA) and Instance-guided Temporal Attention(ITA). ISA computes the correlation between all tokens within one frame, which can gather the context depth features from other human instances or objects. Based on the output of ISA, ITA calculates attention among a group of corresponding tokens in the temporal dimension, which can aggregate depth information of the same instance from different video frames. Since the inputs are whole images, the human instances in images suffer variational scales. To tackle this problem, we propose a cross-scale attention mechanism for IVT. It enables IVT to be more robust to handle the different scales of human instances.

\subsubsection{Instance-guided Spatial Attention}

Instance-guided Spatial Attention (ISA) conducts spatial self-attention within one frame. Here, for the $t$th frame, it is tokenized as instance-guided tokens $\tau_t$. The tokens are sent into ISA and output $\tau^{ISA}_t$, which can be formulated as:
\begin{equation}
\label{eq_aggre}
    \tau^{ISA}_t = A^{ISA}(\tau_t) = FFN(MHSA(\tau_t))
\end{equation}
where $MHSA$ and $FFN$ are multi-head self-attention and feed-forward network, respectively. $A^{ISA}(\cdot)$ means instance-guided spatial attention. Due to the instance-guided tokens, ISA can capture more fine-grained keypoints relationships between the same person and different human instances at the same time.
We compute ISA on each video frame to obtain a sequence of token maps denoted as $\tau^{ISA}=\{\tau^{ISA}_t | t \in [1, T]\}$.

\subsubsection{Instance-guided Temporal Attention}
Temporal information is important for 3D human pose estimation, especially in handling occlusion problems. To capture global depth information from temporal features, we introduce instance-guided temporal attention (ITA) here, which computes the cross-attention on instance-guided tokens from different video frames.

The query, key, and value for ITA are generated from the token maps $\tau^{ISA}$ outputted from ISA.
For a query token $\tau^{ISA}_T[i]$ at the $i$th block in $T$th frame, ITA is computed on the same block places among different frames, and we denote the query, key and value for updating this token as $\mathcal{Q}_T[i]$, $\mathcal{K}_T[i]$ and $\mathcal{V}_T[i]$ respectively, which are formulated as:
\begin{equation}
\begin{aligned}
    \mathcal{Q}_T[i] &= \mathbf{P}_q(\tau^{ISA}_T[i]),\\
    \mathcal{K}_T[i] &= \mathbf{P}_k(Concat_n(\tau^{ISA}_1[i],...,\tau^{ISA}_{T-1}[i],\tau^{ISA}_T[i])),\\
    \mathcal{V}_T[i] &= \mathbf{P}_v(Concat_n(\tau^{ISA}_1[i],...,\tau^{ISA}_{T-1}[i],\tau^{ISA}_T[i]))
\end{aligned}
\end{equation}
where $\mathbf{P}_{q}$, $\mathbf{P}_{k}$, and $\mathbf{P}_{v}$ are linear projection layers for generating the query, key, and value, respectively. $Concat_n$ means concatenating matrixes along sample dimension. $i \in [1, N]$ represents the block index. It is worth to note that, before conducting temporal attention, all the token maps from the video sequence are aligned with optical flow, which is calculated between each pair of adjacent frames in advance. Therefore, the ITA can be computed on the tokens from different frames with the same block index $i$. As shown in Figure~\ref{fig:framework}, the green line represents the corresponding relationship between adjacent frames. Then, for the $i$th token at $T$th frame, the instance-guided temporal attention(ITA) is computed as:
\begin{equation}
    \tau^{ITA}_T[i] = A^{ITA}(\tau^{ISA}_T[i]) = FFN(MHA(\mathcal{Q}_T[i], \mathcal{K}_T[i], \mathcal{V}_T[i]))
\end{equation}
where $A^{ITA}(\cdot)$ means instance-guided temporal attention and it is repeated for all $t \in [1, T]$ and $i \in [1,N]$ for outputting token maps $\tau^{ITA} = \{\tau^{ITA}_t | t \in [1, T]\}$.

Combined with ISA and ITA, the final output of IVT can be formulated as:
\begin{equation}
\label{eq:ivt1}
    \tau^{IVT}_t = A^{ITA}(A^{ISA}(\tau_t)) + \tau_t,\quad t\in [1,T]
\end{equation}

\begin{figure}[!t]
\centering
\includegraphics[width=\columnwidth]{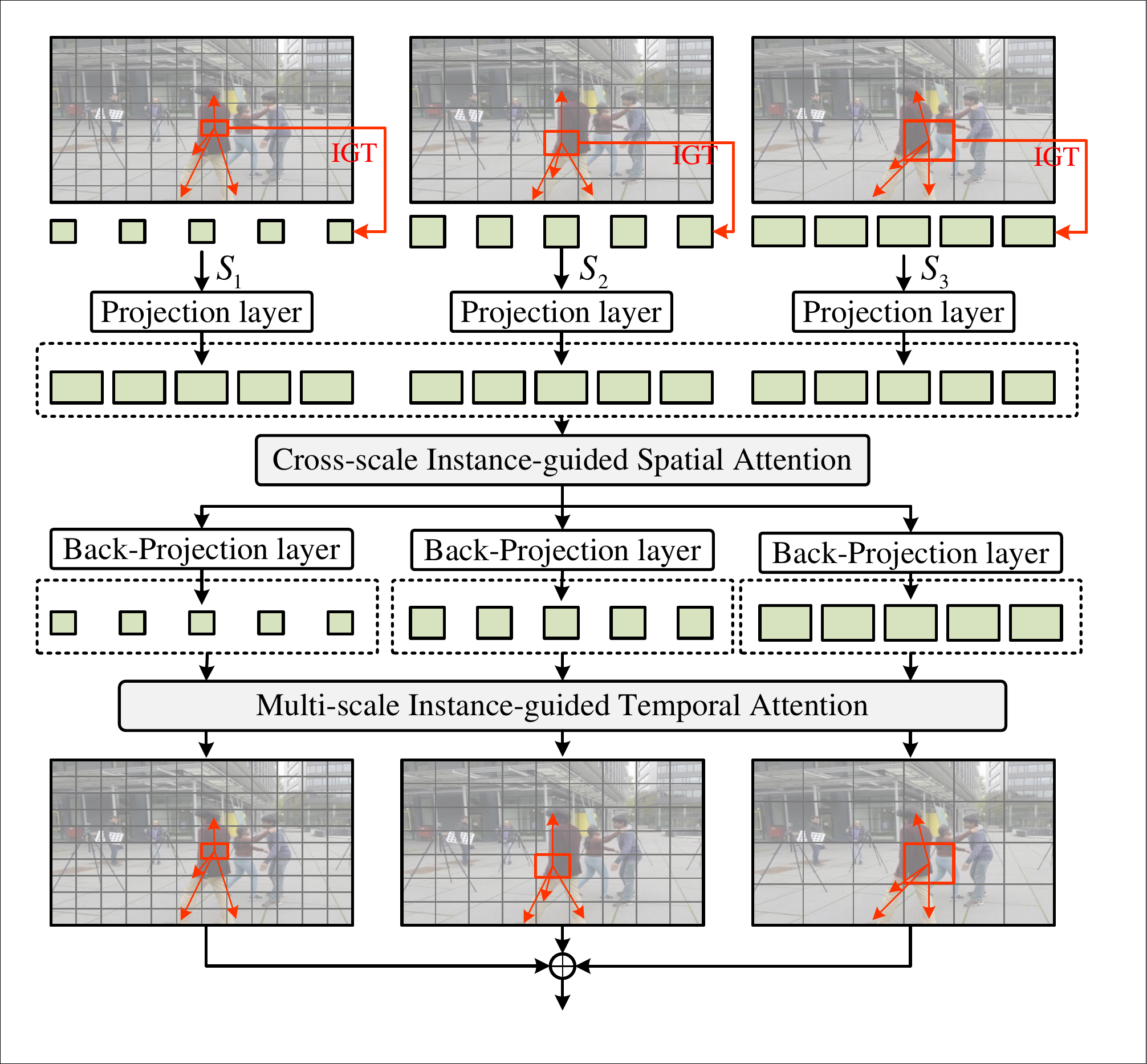}
\vspace{-0.5cm}

\caption{The illustration of the cross-scale instance-guided attention for IVT. $S_1$, $S_2$, and $S_3$ represents three different scales for cross-scale attention. IGT means instance-guided tokenization. The projection layer is a Linear layer to project visual tokens from different scales into the same size. Back-projection is the inverse operation of the projection layer.}
\label{fig:crossattention}
\end{figure}

\subsubsection{Cross-scale Attention}
Scale information also matters in depth estimation because the relative depth of different subjects is correlated with their scales. Therefore, to improve the accuracy of depth estimation, we design a cross-scale attention mechanism for the instance-guided video transformer. This cross-scale mechanism encourages the video transformer to calculate attention between blocks with different scales, which can make the relative depth information from other persons be aggregated better.

Concretely, we split the deep feature map $F_t$ with different block sizes at the instance-guided tokenization stage. In this paper, we apply three block sizes $s \in \{2,4,8\}$. Then, we pass all the tokens from three scales into cross-scale instance-guided spatial attention (CISA), as shown in Figure~\ref{fig:crossattention}. Due to the feature dimension of tokens from different scales being different, we add projection layers before the cross-scale attention for aligning the feature dimension. Meanwhile, back-projection layers are added after cross-scale attention for restoring the feature dimension of those tokens to their original state.

After cross-scale instance-guided spatial attention for each frame, three sequences of token maps with different scales are outputted. Then these three sequences are passed into ITA to perform temporal attention individually. Finally, token maps from different scales are added up into one token map frame by frame. For clearity, we name the above operation as multi-scale instance-guided temporal attention (MITA).

In a nutshell, combined with CISA and MITA, the whole process of IVT in Equation \ref{eq:ivt1} can be recapped as 
\begin{equation}
    \tau^{IVT}_t = A^{MITA}(A^{CISA}(\tau_t)) + \tau_t,\quad t\in [1,T]
\end{equation}
where $A^{CISA}(\cdot)$ denotes cross-scale instance-guided spatial attention and $A^{MITA}(\cdot)$ denotes multi-scale instance-guided temporal attention.

\subsection{Loss Function}
\label{decoding}

For $t$th frame in video, the outputted token map $\tau^{IVT}_t$ are used to learn root keypoints heatmaps $M_{h}$ of size $H\times W$ and 3D offset $M_{o}$ of size $J\times H \times W$ by a convolutional layer. For each point $p$ in $M_{h}$ with high confidence, the corresponding 3D offsets at the $p$ place in $M_{o}$ are extracted to decode a whole 3D pose of size $J\times 3$ for a person, $J$ represents joints number. Finally, NMS is used to remove the superfluous 3D poses. The decoding process is same as previous work~\cite{nie2019single,zhou2019objects}.

During training, we use $L_1$ loss for 3D offsets regression and $L_2$ loss for heatmap learning. Meanwhile, the instance 2D offsets $M^{2D}_o$ are learned with the supervision of ground-truth instance offsets $\hat{M}^{2D}_o$.
The loss function $\mathcal{L}$ is
\begin{equation}
    \mathcal{L} = L_1(M_o, \hat{M}_o) + L_1(M^{2D}_o, \hat{M}^{2D}_o)+ \alpha L_2(M_h, \hat{M}_h),
\end{equation}
where $\hat{M}_o^j$ and $\hat{M}_h$ means the ground-truth of $M_o$ and $M_h$, respectively. $\alpha$ represents a loss weight.

\begin{table*}[!t]
\renewcommand\tabcolsep{2pt}
\centering
 \caption{Quantitative comparison with state-of-the-art methods on Human3.6M under Protocol 1 (MPJPE) and Protocol 2 (PA-MPJPE). $f$ denotes the number of input frames used in each method, and $*$ represents a Transformer-based model. \textbf{Bold} indicates the best and \underline{underline} indicates the second best.}
 
 \vspace{-0.3cm}
 \resizebox{\linewidth}{!}{
 \begin{tabular}{lccccccccccccccccc}
\hline

\hline
\multicolumn{1}{c|}{\textbf{Protocol 1}}       &  \multicolumn{1}{c|}{}      & Dir. & Disc. & Eat. & Greet & Phone & Photo & Pose & Purch. & Sit  & SitD. & Somke & Wait & WalkD. & Walk & \multicolumn{1}{c|}{WalkT.} & Avg. \\ 
\hline
\multicolumn{1}{l|}{Dabral \etal \cite{dabral2018learning} (f=243)}   & \multicolumn{1}{c|}{ECCV'18}      & 44.8 & 50.4  & 44.7 & 49.0  & 52.9  & 61.4  & 43.5 & 45.5   & 63.1 & 87.3  & 51.7  & 48.5 & 52.2   & 37.6 & \multicolumn{1}{c|}{41.9}   & 52.1 \\

\multicolumn{1}{l|}{Cai \etal \cite{cai2019exploiting} ($f=7$)}    & \multicolumn{1}{c|}{ICCV'19 }    & 44.6 & 47.4  & 45.6 & 48.8  & 50.8  & 59.0  & 47.2 & 43.9   & 57.9 & 61.9  & 49.7  & 46.6 & 51.3   & 37.1 & \multicolumn{1}{c|}{39.4}   & 48.8    \\

\multicolumn{1}{l|}{Pavllo \etal \cite{pavllo20193d} ($f=243$)}  & \multicolumn{1}{c|}{CVPR'19 } & 45.2 & 46.7  & 43.3 & 45.6  & 48.1  & 55.1  & 44.6 & 44.3   & 57.3 & 65.8  & 47.1  & 44.0 & 49.0   & 32.8 & \multicolumn{1}{c|}{33.9}   & 46.8    \\

\multicolumn{1}{l|}{Lin \etal \cite{lin2019trajectory} ($f=50$)}   & \multicolumn{1}{c|}{BMVC'19}     & 42.5 & 44.8  & 42.6 & 44.2  & 48.5  & 57.1  & 52.6 & 41.4   & 56.5 & 64.5  & 47.4  & 43.0 & 48.1   & 33.0 & \multicolumn{1}{c|}{35.1}   & 46.6    \\
\multicolumn{1}{l|}{Yeh \etal \cite{yeh2019chirality} ($f=243$)}       & \multicolumn{1}{c|}{NeurIPS'19}       & 44.8 & 46.1  & 43.3 & 46.4  & 49.0  & 55.2  & 44.6 & 44.0   & 58.3 & 62.7  & 47.1  & 43.9 & 48.6   & 32.7 & \multicolumn{1}{c|}{33.3}   & 46.7    \\

\multicolumn{1}{l|}{Liu \etal \cite{liu2020attention} ($f=243$)}   & \multicolumn{1}{c|}{CVPR'20}  &   41.8 & 44.8  & 41.1 & 44.9  & 47.4  & 54.1  & 43.4 & 42.2   & 56.2 & 63.6  & \underline{45.3}  & 43.5 & 45.3   & 31.3 & \multicolumn{1}{c|}{32.2}   & 45.1    \\

\multicolumn{1}{l|}{Zeng \etal \cite{zeng2020srnet}  ($f=243$)} & \multicolumn{1}{c|}{ECCV'20} & 46.6 & 47.1  & 43.9 &\underline{ 41.6} & 45.8 & \underline{49.6} & 46.5 & \textbf{40.0} & 53.4 & 61.1  & 46.1  & 42.6 & \underline{43.1} & 31.5 & \multicolumn{1}{c|}{32.6}  & 44.8   \\

\multicolumn{1}{l|}{Wang \etal \cite{wang2020motion} ($f=96$)}    & \multicolumn{1}{c|}{ECCV'20}  & \underline{41.3} & 43.9  & 44.0 & 42.2  & 48.0  & 57.1  & 42.2 & 43.2   & 57.3 & 61.3  & 47.0  & 43.5 & 47.0  & 32.6 & \multicolumn{1}{c|}{31.8}   & 45.6    \\

\multicolumn{1}{l|}{Chen \etal \cite{chen2021anatomy} ($f=81$)}  & \multicolumn{1}{c|}{TCSVT'21}  & 42.1 & \underline{43.8} & 41.0 & 43.8  & \underline{46.1}  & 53.5  & 42.4 & 43.1   & 53.9 & {60.5}  & 45.7  &\underline{ 42.1} & 46.2   & {32.2} & \multicolumn{1}{c|}{33.8}   & 44.6    \\
\hline
\multicolumn{1}{l|}{Lin \etal \cite{lin2021end} ($f=1$)$^*$}    & \multicolumn{1}{c|}{CVPR'21}    & -    & -     & -    & -     & -     & -     & -    & -      & -    & -     & -     & -    & -      & -    & \multicolumn{1}{c|}{-}      & 54.0 \\
\multicolumn{1}{l|}{Liu \etal \cite{liu2020attention} ($f=243$)$^*$}    & \multicolumn{1}{c|}{ICRA'21} & 43.3 & 46.1 & 40.9 & 44.6 & 46.6 & 54.0 & 44.1 & 42.9 & 55.3  & \textbf{57.9} & 45.8 & 43.4 & 47.3  & 30.4 & \multicolumn{1}{c|}{\textbf{30.3}} & 44.9 \\

\multicolumn{1}{l|}{Zheng \etal \cite{zheng20213d} ($f=81$)$^*$} & \multicolumn{1}{c|}{ICCV'21}  & 41.5 & 44.8  & \underline{39.8} & 42.5  & 46.5  & 51.6  & \underline{42.1} & \underline{42.0}   & \underline{53.3} & 60.7  & 45.5  & 43.3 & 46.1   & \textbf{31.8} & \multicolumn{1}{c|}{\underline{32.2}}   &\underline{44.3}    \\ 
\hline
\multicolumn{1}{l|}{\textbf{Ours (IVT)} ($f=5$)$^*$} & \multicolumn{1}{c|}{}  &  \textbf{36.5} & \textbf{40.1} &  \textbf{38.4} & \textbf{40.7} & \textbf{42.6} & \textbf{42.8} & \textbf{30.1} & 43.4 & \textbf{46.1} & \underline{58.0} & \textbf{40.2} & \textbf{37.1} & \textbf{40.8} & \underline{32.1} & \multicolumn{1}{c|}{33.5} & \textbf{40.2}  \\
\hline

\hline
\multicolumn{1}{c|}{\textbf{Protocol 2}}       & \multicolumn{1}{c|}{}        & Dir. & Disc. & Eat. & Greet & Phone & Photo & Pose & Purch. & Sit  & SitD. & Somke & Wait & WalkD. & Walk & \multicolumn{1}{c|}{WalkT.} & Avg. \\ 
\hline

\multicolumn{1}{l|}{Hossain \etal \cite{hossain2018exploiting} ($f=243$)}     & \multicolumn{1}{c|}{ ECCV'18}     & 35.7   & 39.3  & 44.6 & 43.0  & 47.2  & 54.0  & 38.3 & 37.5   & 51.6 & 61.3  & 46.5  & 41.4 & 47.3   & 34.2 & \multicolumn{1}{c|}{39.4}   & 44.1    \\

\multicolumn{1}{l|}{Cai \etal \cite{cai2019exploiting} ($f=7$)}   & \multicolumn{1}{c|}{ICCV'19}     & 35.7 & 37.8  & 36.9 & 40.7  & 39.6  & 45.2  & 37.4 & 34.5   & 46.9 & 50.1  & 40.5  & 36.1 & 41.0   & 29.6 & \multicolumn{1}{c|}{32.3}   & 39.0    \\

\multicolumn{1}{l|}{Lin \etal \cite{lin2019trajectory} ($f=50$)}    & \multicolumn{1}{c|}{BMVC'19}    & 32.5 & 35.3  & 34.3 & 36.2  & 37.8  & 43.0  & 33.0 & 32.2   & 45.7 & 51.8  & 38.4  & 32.8 & 37.5   & 25.8 & \multicolumn{1}{c|}{28.9}   & 36.8    \\

\multicolumn{1}{l|}{Pavllo \etal \cite{pavllo20193d} ($f=243$)}  & \multicolumn{1}{c|}{CVPR'19 } & 34.1 & 36.1  & 34.4 & 37.2  & 36.4  & 42.2  & 34.4 & 33.6   & 45.0 & 52.5  & 37.4  & 33.8 & 37.8   & 25.6 & \multicolumn{1}{c|}{27.3}   & 36.5    \\

\multicolumn{1}{l|}{Liu \etal \cite{liu2020attention} ($f=243$)}   & \multicolumn{1}{c|}{CVPR'20 }  & \underline{32.3} &  35.2  & 33.3 & 35.8  & 35.9  & 41.5  & 33.2 & 32.7   & 44.6 & 50.9  & 37.0  & \underline{32.5} & 37.0   &  \underline{25.2} & \multicolumn{1}{c|}{27.2}   & 35.6    \\  		

\multicolumn{1}{l|}{Wang \etal \cite{wang2020motion} ($f=96$)}   & \multicolumn{1}{c|}{ECCV'20}   & 32.9 &  35.2  & 35.6 & \underline{34.4}  & 36.4  & 42.7  & \underline{31.2} & 32.5   & 45.6 & 50.2  & 37.3  & 32.8 & 36.3   & 26.0 & \multicolumn{1}{c|}{\textbf{23.9}}   & 35.5    \\

\multicolumn{1}{l|}{Chen \etal \cite{chen2021anatomy} ($f=81$)}  & \multicolumn{1}{c|}{TCSVT'21 } & 33.1 & 35.3  & 33.4 & 35.9  & 36.1  & 41.7  & 32.8 & 33.3   & \underline{42.6} & 49.4  & 37.0  & 32.7 & 36.5   & 25.5 & \multicolumn{1}{c|}{27.9}   & 35.6    \\ 
\hline
\multicolumn{1}{l|}{Liu \etal \cite{liu2020attention} ($f=243$)$^*$}    & \multicolumn{1}{c|}{ICRA'21} & 32.7 & 36.2 & 33.4 & 36.5 & 36.0 & 41.5 & 33.6 & 33.1 & 44.1 & 46.8 & 36.7 & 33.1 & 35.8  & 24.2 & \multicolumn{1}{c|}{24.8} & 35.2 \\

\multicolumn{1}{l|}{Zheng \etal \cite{zheng20213d} ($f=81$)$^*$}        & \multicolumn{1}{c|}{ICCV'21}    & 32.5 & \underline{34.8}  & \underline{32.6} & 34.6  & \underline{35.3}  & \underline{39.5}  & 32.1 & \underline{32.0}   & 42.8 & \underline{48.5}  & \underline{34.8}  & \textbf{32.4} & \underline{35.3}   & \textbf{24.5} & \multicolumn{1}{c|}{26.0} & \underline{34.6} \\
\hline
\multicolumn{1}{l|}{\textbf{Ours (IVT)} ($f=5$)$^*$} & \multicolumn{1}{c|}{}  & \textbf{27.0} & \textbf{24.8}  & \textbf{32.2} & \textbf{30.1}  & \textbf{27.8}  &  \textbf{32.1} & \textbf{22.3} & \textbf{28.7} & \textbf{30.7} & \textbf{24.4}  & \textbf{32.7}  & 37.8 & \textbf{21.9} & 31.1 & \multicolumn{1}{c|}{\underline{24.7}} & \textbf{28.5} \\
\hline

\hline
\end{tabular}
}
\label{tab:h36m}
\end{table*}

\section{Experiments}
In this section, we elaborate the experiment results of IVT. We firstly introduce the implemental details of IVT, and then report results and compare with SOTA methods on three widely-used datasets: Human3.6M, 3DPW, and CMU Panoptic. All ablation studies are based on CMU Panoptic dataset. Meanwhile, some visualization results on Human3.6M are given for presenting the superiority of IVT in an intuitionistic way.

\subsection{Implemental Details}
We use HRNet-32~\cite{wang2020deep} pre-trained on 2D pose estimation dataset COCO~\cite{lin2014microsoft} as the backbone network of IVT and SPyNet~\cite{ranjan2017optical} as the motion estimation network between different frames. In our experiments, IVT is stacked with 3 layers, and it is trained on 8 V100 GPUs with a batch size of 4 sequences/GPU, while the sequence length is 5 frames and the input size is $512\times 512$. The total training epochs is 50. Adam optimizer is adopted and the initial learning rate is 5e-4, which decreases 10× at 30 and 40 epochs. The loss weight $\alpha$ equals 10 during training.

\subsection{Datasets and Metrics}

\subsubsection{Human3.6M dataset}
Human3.6~\cite{ionescu2013human3} is the largest indoor benchmark for single-person 3D pose estimation, which includes 7 subjects that performing 15 actions. Following the previous works~\cite{zeng2020srnet,moon2019camera,zheng20213d,lin2021end}, we use two protocols for evaluation. 
For \textbf{Protocol 1}, IVT is trained on the subjects S1, S5, S6, S7, and S8, and tested on the subjects S9 and S11 by using Mean-Per-Joint-Position-Error (MPJPE), which measures the Euclidean distances between the ground truth joints and the predicted joints. 
For \textbf{Protocol 2}, subjects S1, S5, S7, S8, and S9 are used for training, and S11 is used for testing by PA-MPJPE. It calculates the Euclidean distance between predicted and ground-truth 3D joint coordinates after root joint alignment and further rigid alignment by Procrustes analysis~\cite{gower1975generalized}.

\subsubsection{3DPW dataset} 
3DPW~\cite{von2018recovering} is a multi-person outdoor 3D pose estimation dataset, which contains 22K images for training and 35K images for testing. Following the previous works~\cite{kocabas2020vibe,lin2021end,choi2021beyond}, we train IVT on the training set and evaluate IVT on the testing set in PA-MPJPE.

\subsubsection{CMU Panoptic dataset}
CMU Panoptic~\cite{joo2017panoptic} is a larger-scale multi-person dataset, captured by multiple cameras. Following the settings of previous works~\cite{wang2020hmor,zhen2020smap}, we use 160K images from different videos as the training set and the videos from two cameras (16, 30) as the testing set. For comparison, MPJPE is used for evaluation. 

\subsection{Comparison with SOTA Methods}

\subsubsection{Results on Human3.6M Dataset}
The comparisons with state-of-the-art methods on the Human3.6M dataset are shown in Table \ref{tab:h36m}. Our IVT with $f=5$ achieves new state-of-the-art results with an MPJPE of 41.3mm and a PA-MPJPE of 28.5mm in Protocol 1 and Protocol 2, respectively. The relative gains are 6.8\% and 17.6\%, respectively. The results demonstrate the effectiveness of the proposed IVT. 
Compared with other transformer-based methods~\cite{lin2021end,zheng20213d,liu2021graph}, IVT outperforms them. Even the PoseFormer~\cite{zheng20213d} is based on a frame number 81, IVT with $f=5$ obtains better results since the PoseFormer loses the visual depth feature in the process of temporal modeling.

\begin{figure*}[!t]
\centering
\includegraphics[width=\textwidth]{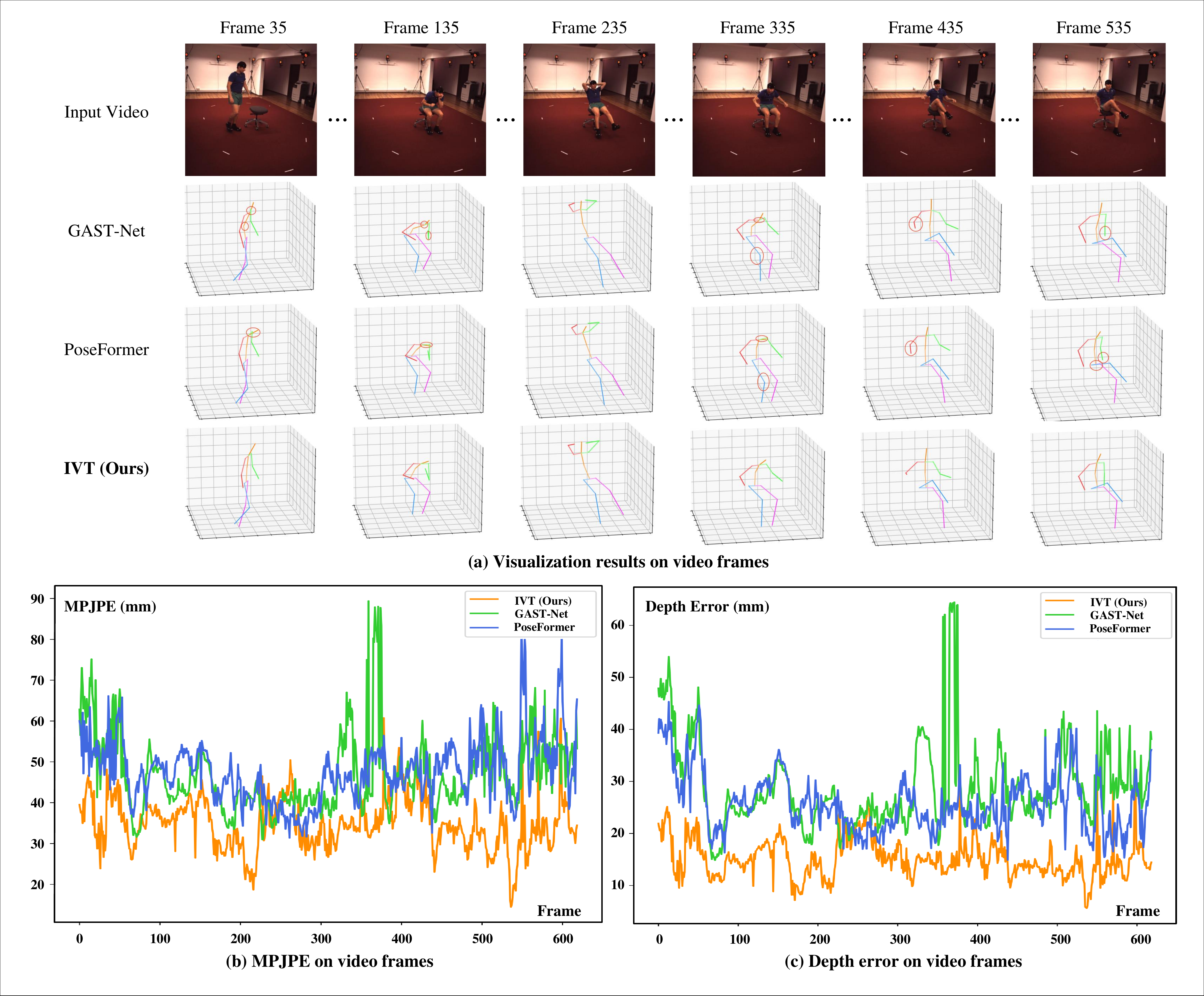}
\vspace{-0.5cm}
\caption{Qualitative comparison between IVT and the SOTA video methods (GAST-Net~\cite{liu2021graph} and PoseFormer~\cite{zheng20213d}) on video from Human3.6M dataset. (a) visualization results (Red circle denotes the wrong prediction). (b) The curve of MPJPE-Frame shows the MPJPE on each frame. (c) The curve of Depth Error-Frame shows the MPJPE in depth dimension on each frame. The depth prediction by IVT is better than GAST-Net and PoseFormer. Best viewed in color.}
\label{fig:video_ana}
\end{figure*}

We also give a fine-grained analysis of videos from the Human3.6M dataset in Figure \ref{fig:video_ana}. 
As shown in Figure \ref{fig:video_ana} (a), given the video inputs, GAST-Net~\cite{liu2021graph} and PoseFormer~\cite{zheng20213d} fail on these hard cases with complex postures or occlusions, but our IVT performs well on these cases since the captured temporal depth context. The red circle denotes the wrong pose predicted by GAST-Net and PoseFormer.

As shown in Figure \ref{fig:video_ana} (b) and (c), the MPJPE of IVT is lower than GAST-Net and PoseFormer, and the depth error of IVT is lower than GAST-Net and PoseFormer. It shows that the improvements of IVT mainly benefit from better depth prediction. Combined with Figure \ref{fig:video_ana} (a), (b), and (c), we can found that GAST-Net, PoseFormer, and IVT have similar results on frame 235. The MPJPE and depth error show the similar results at frame 235 since the human poses in this period are clear.

\subsubsection{Results on 3DPW Dataset}
The comparisons with state-of-the-art methods on the 3DPW dataset are shown in Table \ref{tab:3dpw}. In the multi-person 3DPW dataset, IVT outperforms previous video-based methods and achieves 46.0mm in PA-MPJPE. Compared with METRO~\cite{lin2021end}, the relative gain is 6\% in PA-MPJPE. These results verify the effectiveness and generalization ability of the proposed IVT since 3DPW is an in-the-wild dataset.

\begin{table}[]
\renewcommand\tabcolsep{5pt}
\centering
 \caption{Quantitative comparison with state-of-the-art methods on multi-person 3D human pose estimation dataset (3DPW) in PA-MPJPE. Frame denotes the number of input frames used in each method. $*$ denotes transformer-based methods. Lower is better.}
 \vspace{-0.3cm}
 \resizebox{\linewidth}{!}{
 \begin{tabular}{l|c|c|c}
 \hline
 
 \hline
 Methods  & &Frames & PA-MPJPE $\downarrow$ \\
 \hline
  Doersch \etal \cite{doersch2019sim2real}& NeurIPS'19 & 31 & 74.4 \\
  Kanazawa \etal \cite{kanazawa2019learning}& CVPR'19 &10 & 72.6 \\
  Cheng \etal \cite{cheng20203d} & AAAI'20 & >90 & 71.8\\
  Sun \etal \cite{sun2019human}& ICCV'19 &45 & 69.5 \\
  Kolotouros \etal \cite{kolotouros2019learning} & ICCV'19 & 1 & 59.2\\
  Kocabas \etal \cite{kocabas2020vibe} & CVPR'20 &16 & 57.6\\
  Luo \etal \cite{luo20203d} & ACCV'20 & 90 & 54.7\\
  Cheng \etal \cite{cheng2021monocular}& CVPR'21 & >90 & 62.9\\
  Choi \etal \cite{choi2021beyond} & CVPR'21 & 16 & 52.7\\
  \hline
   \textbf{Ours (IVT)}* & & 5 & \textbf{46.0} \\
 
 \hline
 
 \hline
 
 \end{tabular}
 }
 \label{tab:3dpw}
\end{table}

\subsubsection{Results on CMU Panoptic Dataset}
The comparisons with SOTA methods on CMU Panoptic dataset are shown in Table \ref{tab_cmu}. 
It shows that, IVT obtains 48.4mm in MPJPE and achieves a new state-of-the-art result with a relative gain of 10\% compared with the the DAS~\cite{wang2022distribution}. The results on CMU Panoptic dataset further demonstrate the effectiveness of the proposed IVT framework.

\begin{table}[]
\renewcommand\tabcolsep{6pt}
\centering
 \caption{Comparison with SOTA methods on multi-person 3D human pose estimation dataset (CMU Panoptic) in MPJPE. ${\dagger}$ means an extra refining network is used. Lower is better.}
 \vspace{-0.3cm}
 \begin{tabular}{l|c|c}
 \hline
 
 \hline
 Methods &  & MPJPE (mm)~$\downarrow$ \\
 \hline
 SFB~\cite{zanfir2018monocular} & CVPR'18 &  153.4\\
 PoseNet~\cite{moon2019camera} & ICCV'19 &  87.6\\
 MubyNet~\cite{zanfir2018deep} & NeurIPS'18  & 78.1\\
 SMAP~\cite{zhen2020smap} & ECCV'20  & 73.1\\
 LoCO~\cite{fabbri2020compressed} & CVPR'20  & 69.0\\
 SMAP$^{\dagger}$~\cite{zhen2020smap} & ECCV'20  & 61.8\\
 DAS~\cite{wang2022distribution} & CVPR'22  & 53.8\\
 \hline
 \textbf{Ours(IVT)} &  & \textbf{48.4}\\

 \hline
 
 \hline
 \end{tabular}
 \label{tab_cmu}
\end{table}

\begin{table}[]
\renewcommand\tabcolsep{2pt}
\centering
 \caption{Ablation study of IVT on CMU Panoptic. SA means traditional spatial attention. ISA means using instance-guided spatial attention in IVT. ITA means using instance-guided temporal attention in IVT. CISA represents cross-scale ISA. MITA means multi-scale ITA. Flops are computed on two input images with a size of $512\times 512$.}
 \vspace{-0.3cm}
 \resizebox{\linewidth}{!}{
 \begin{tabular}{l|l|c|c|c|c}
 \hline
 
 \hline
 Methods  & Feature & Params (M) & Flops (T) & MPJPE (mm)$\downarrow$ & $\Delta$ \\
 \hline
 Baseline & + SA & 32.96 & 0.127 & 52.0 & -\\
 IVT & + ISA & 34.88 & 0.123 & 50.8 & $\downarrow 2.3\% $\\
 IVT & + ISA + ITA & 35.81 & 0.126 & 49.5 &$\downarrow 2.6\% $\\
 IVT & + CISA + MITA & 40.85 & 0.134 & \textbf{48.4} &$\downarrow 2.3\% $\\
 \hline
 
 \hline
 \end{tabular}
 }
 \label{tab_ab_cmu}
\end{table}

\begin{table}[]
\renewcommand\tabcolsep{7pt}
\centering
 \caption{Ablation study of IVT (ISA-ITA) on frame numbers on CMU Panoptic dataset. Note that each video is sampled with a sampling rate of 5 frames. Thus, the temporal receptive field is $r=f\times 5$, where $f$ means the used frame number.}
 \vspace{-0.3cm}
 \begin{tabular}{c|c|c|c|c|c}
 \hline
 
 \hline
 Frames & 1 & 3 & 5 & 7 & 9\\
 \hline
 MPJPE (mm) & 51.8 & 50.3 & 49.5 & 49.5 & 50.0\\
 \hline
 
 \hline
 \end{tabular}
 \label{tab_abt}
\end{table}

\subsection{Ablation Study}
In this section, we verify the effectiveness of the proposed ISA, ITA, and cross-scale attention mechanism in instance-guided video transformer (IVT). Based on IVT, we also study the influence of frame numbers on video transformer. Then, we compare the parameters and computational costs of different types of IVT.

\subsubsection{Effectiveness of proposed attention mechanisms}
We conduct the ablation study on CMU Panoptic dataset to verify the effectiveness of proposed modules in the instance-guided video transformer. 

First of all, to verify the different types of attention mechanisms, we build an end-to-end multi-person 3D pose estimation baseline with traditional simple spatial attention as \cite{dosovitskiy2020image,bertasius2021space}, noted as SA. As shown in Table \ref{tab_ab_cmu}, SA achieves 52.0mm in MPJPE. Combined with the instance-guided attention, IVT with ISA obtains 50.8mm in MPJPE and achieves a relative gain of 2.3\%. Compared with only using ISA,  the ITA obtains 49.5mm in MPJPE and achieves a relative gain of 2.6\%. Besides, the cross-scale attention mechanism brings a relative gain of 2.3\% and achieves 48.4mm in MPJPE. These results show that the proposed ISA, ITA, and cross-scale attention mechanism are useful for 3D human pose estimation.

\subsubsection{The ablation study on frame number}
To explore the influence of frame numbers for IVT, we conduct the ablation study of frame numbers based on IVT with ITA. As shown in Table \ref{tab_abt}, the frames number means the used frames, but the truth receptive field on time is $f\times 5$ since the sampling rate of the video is 5 frames. For example, $f=5$ means we used 5 frames but the interval is 25 from the first frame to the last frame. As shown in Table \ref{tab_abt}, given more frames from 1 to 5, the performance of IVT improves to 49.5mm from 50.8mm. But with the frame number increasing to 9, the performance of IVT decreases to 50.00mm. The result shows that long-term frames could damage the performance of IVT since the long-term motion is hard to estimate.

\subsubsection{Parameters and computational costs}

The comparisons of different attention mechanisms on parameters and computational costs are shown in Table \ref{tab_ab_cmu}. Compared with the baseline with spatial attention, instance-guided attention (ISA) obtains a relative gain of 2.3\% by adding 1.92MB parameters, while the flops of ISA decrease since the instance-aware tokens are based on image blocks. Even for the IVT with CISA and ITA, the increasing parameters and the computational costs are acceptable.

\section{Conclusions}
In this paper, we propose a novel end-to-end instance-guided video transformer (IVT) for video 3D human pose estimation to capture global depth context.
To capture the depth context in both spatial and temporal dimensions, we propose instance-guided spatial attention (ISA) and instance-guided temporal attention (ITA) mechanisms. To further tackle the variational human scales in video, we propose cross-scale attention for IVT. Combined with ISA, ITA, and cross-scale attention, IVT outperforms state-of-the-art methods on three widely-used 3D human pose estimation datasets.

\section{Acknowledgement}
This work was supported by the Scientific and Technological Innovation of Shunde Graduate School of University of Science and Technology Beijing (No. BK20AE004 and No.BK19CE017).

\clearpage
\bibliographystyle{ACM-Reference-Format}
\bibliography{sample-base}

\end{document}